\documentclass[conference]{IEEEtran}

\usepackage{algorithm2e}
\usepackage{grffile} % so that file-0.0.pdf is recognized
\usepackage{xcolor}
\usepackage{supertabular} % can split table over two columns
\usepackage{multirow}
\usepackage{listings}

\usepackage[hyphens]{url}
\usepackage{hyperref}
\hypersetup{breaklinks=true}

\usepackage{subcaption}

\usepackage{mathtools} 
\usepackage{cleveref}
\usepackage{enumitem}
\usepackage{lipsum}
\setlist[itemize]{leftmargin=*}

\def\BibTeX{{\rm B\kern-.05em{\sc i\kern-.025em b}\kern-.08em
    T\kern-.1667em\lower.7ex\hbox{E}\kern-.125emX}}

\begin{document}

%%%%%%%%%%%%%%%%%%%%%%%%%%%%%%%%%%%%%%%%%%%%%%%%%%%%%%%%%%%%%%%%%%%%%%%
\title{Financial Crime \& Fraud Detection Using Graph Computing: Application Considerations \& Outlook}
%%%%%%%%%%%%%%%%%%%%%%%%%%%%%%%%%%%%%%%%%%%%%%%%%%%%%%%%%%%%%%%%%%

% \author{Hongda Shen}
% \affiliation{
%   \institution{University of Alabama in Huntsville}
% }
% \email{hs0017@alumni.uah.edu}

% \author{Haojie Yu}
% \affiliation{
%   \institution{Georgia Institute of Technology}
% }
% \email{haojie.yu@gatech.edu}

% \author{Eren Kursun}
% \affiliation{
%   \institution{Columbia University}
% }
% \email{ek2925@columbia.edu}

\author{
\IEEEauthorblockN{Eren Kurshan}
\IEEEauthorblockA{\textit{Columbia University} \\
ek2925@columbia.edu}
\and
\IEEEauthorblockN{Hongda Shen}
\IEEEauthorblockA{\textit{University of Alabama in Huntsville} \\
hs0017@alumni.uah.edu}
\and
\IEEEauthorblockN{Haojie Yu}
\IEEEauthorblockA{\textit{Georgia Institute of Technology} \\
hyu406@gatech.edu}
}
\maketitle

\begin{abstract}
In recent years, the unprecedented growth in digital payments fueled consequential changes in fraud and financial crimes. In this new landscape, traditional fraud detection approaches such as rule-based engines have largely become ineffective. AI and machine learning solutions using graph computing principles have gained significant interest. Graph neural networks and emerging adaptive solutions provide compelling opportunities for the future of fraud and financial crime detection. However, implementing the graph-based solutions in financial transaction processing systems has brought numerous obstacles and application considerations to light. In this paper, we overview the latest trends in the financial crimes landscape and discuss the implementation difficulties current and emerging graph solutions face. We argue that the application demands and implementation challenges provide key insights in developing effective solutions.
\end{abstract}

% \keywords{Artificial Intelligence, Machine Learning, Fraud Detection, Financial Crime Detection, Money Laundering, Graph Computing, Algorithms}
% \settopmatter{printfolios=true} % Page numbers


\section{Introduction}
\label{sec:intro}
Digital payments have experienced an unparalleled growth in the past decade \cite{McKinsey19},\cite{Fed19},\cite{BostonFed}. In 2019 alone, 743 million transactions, valued at \$187 billion were processed through the Zelle digital payment network alone. This translates to a 57\% year-to-year growth in the total transaction amounts and 72\% increase in the transaction volumes \cite{zelle}. Between 2013-2018, mobile payments grew by over 120\% (in compounded annual growth rate) in China  \cite{McKinsey19}. Similarly in India digital payments volume rose by 61\% over the past 5 years \cite{ReserveBankIndia}.

Globally, mobile banking and digital payments have provided billions of people the opportunity to access financial services. Furthermore, they have delivered practical benefits to individual consumers, businesses and financial service providers, such as time savings, speed, ease of use, lower transaction costs and the ability to scale \cite{Visa}.  On the other hand, criminal schemes have rapidly evolved to benefit from the new fast-moving digital payments landscape. Traditionally, fraud and financial crime detection relied on a large number of rules and static thresholds to flag suspicious transactions (e.g. amounts larger than \$10,000). In recent years, such manual and rule based techniques have become ineffective as fraudsters rapidly figure out the static rules and bypass them. 

Global financial crime volume was estimated to be around \$1.4-\$3.5 trillion per year according to the latest industry reports \cite{EY}. Within this volume, money laundering is estimated to be around 2-5\% of the global GDP (up to 1.87 trillion EUR) \cite{Deloitte}. 3.2 million fraud records were filed through the U.S. Federal Trade Commission System in 2019 alone; indicating a 53\% increase from 2018 \cite{FTC}. The cost to fight and recover from fraud has also increased by over 30\% since 2016 \cite{LexisNexis}. In addition to the increases in the fraud cases, fraud schemes also changed considerably. In the past few years, with the general adoption of the EMV credit and debit cards, in-person card fraud has declined sharply \cite{Beyond_EMV}. Yet, a substantial surge has been reported in the attempted online fraud \cite{transunion}. As of 2020, online credit card fraud cases rose due to the global pandemic and the corresponding increases in the digital transaction volumes \cite{WSJ}. 

Following these trends, AI and machine learning solutions have gained considerable interest from compliance and risk management functions \cite{WEF}. Graph algorithms and databases have long been considered important tools in fraud detection\cite{Mckinsey}. Numerous studies have demonstrated the effective use of anomaly detection, network flow and sub-graph based analysis \cite{DataMining,Fraud_Graph_Survey}.  Lately, graph neural networks have gained interest \cite{GraphML_early}, \cite{Scarselli}, \cite{Scarselli2},\cite{GraphML}. Prior to the financial services applications, graph neural networks have been implemented in a wide range of industries. Their broader potential to improve generalizations and relational reasoning has been highlighted by a number of studies \cite{GNN,google, GNN_Survey}. In financial crime and fraud detection they provide enhanced performance and flexibility \cite{GNN_Survey}. 

Implementing graph-based solutions in real-life transaction processing and crime detection systems brings new challenges to light. In this paper, we take a practical look into the use of graph computing in financial crime detection applications. We highlight the difficulties development organizations face in building and deploying graph-based solutions in financial transaction processing systems. Both financial crime detection and graph computing are very broad and rapidly evolving fields. The purpose of this paper is not to provide a comprehensive overview of these fields (as it should be understood that both will change quickly), but to highlight the general pain-points and overarching practical implementation considerations in order to develop more effective solutions.

The paper is organized as follows: \Cref{sec:overview} is a high-level overview of the recent trends in financial crimes; \Cref{sec:Techniques} discusses highlights from the state-of-the-art research in graph-based solutions;  \Cref{sec:Considerations} reviews the application considerations; and finally \Cref{sec:conclusion} provides conclusion discussions and outlook.  

%%%%%%%%%%%%%%%%%%%%%%%%%%%%%%%%%%%%%%%%%%%%%
\section{Background}
\label{sec:overview}
\subsection{Fraud and Financial Crime Trends}
%\textit{First party fraud} is when an individual or group misrepresents the information or identity during a financial service application. \textit{Second party} when the individual knowingly participates in sharing information to commit fraud. \textit{Third party} is when an individual or group use another individuals information or identity without their knowledge. 
This section provides a high-level overview of the latest trends in financial crimes. Criminal schemes have been going through a transformation lately \cite{IC3_Report},\cite{FED_Survey}. According to the recent industry surveys \cite{KPMG}, almost all fraud types have seen serious increases, with a few exceptions such as mortgage fraud. External fraud has been rising, both in terms of volume and total transaction amounts by 61\% and 59\% respectively.  

%\vspace{8pt}
\subsubsection{Payments Fraud} 
%\noindent \textit{Payment Fraud}
Payments fraud covers fraud in numerous payment channels including credit and debit card transactions, ATM, person-to-person (P2P) transactions, wire, automated clearing house transactions, online payments, automated bill payments, checks and deposits. In recent years, pervasive increases have been seen in fraud across all payment channels, with the biggest increases in digital transactions \cite{FED_Survey}. In addition, frequent cross overs among payment channels and fraud types have been reported. As an example, according to the World Bank \cite{worldbank_P2P} criminals increasingly utilize emerging mobile payment channels for money laundering.
%\vspace{8pt}
\subsubsection{Identity Theft} 
%\noindent \textit{Identity Theft}
ID theft schemes use an ever-changing list of tactics ranging from ATM skimming devices to phishing, smishing, dumpster diving, and compromised wireless networks. Lately, identity theft has become one of the top fraud types in the Federal Trade Commission criminal filings \cite{FTC}. Following the identity theft itself, perpetrators typically use the compromised information across multiple channels. Credit card fraud was the top fraud type for such downstream fraud in 2019, with over 200K cases filed. New credit account fraud grew by 88\% during the same period \cite{FTC}. Identity theft and new account frauds cause more financial damage compared to the other payment fraud types due to the time it takes to detect them. 

%\vspace{8pt}
\subsubsection{Financial and Elderly Scams} 
%\noindent \textit{Financial Scams}
Financial scams have become one of the top concerns in the fraud landscape \cite{FTC}. These crimes use continuously evolving tactics such as phone scams, elderly scams (such as grandparent scam), technology support scams, charity and lottery scams, ticket scams etc. \cite{SEC}. Financial scams usually tie to identity theft and account takeover. They are followed by fraud in one or more payment channels.

\subsubsection{Account Takeover} 
%\noindent \textit{Account Takeover}
Account takeover fraud occurs when perpetrators gain access to a victims account illegally. During this process, criminals typically change the account login credentials and contact information, so that the victim is not able to access the account. They eventually drain the funds through one or more payment channels. ATO has strong ties to cybersecurity as perpetrators regularly use mass data breaches, SIM hijacking, compromised devices and networks to fuel their attacks. Similar to identity theft, ATO provides a gateway to numerous downstream fraud types. It has experienced an 78\% increase in 2019 alone \cite{Javelin}.  

%\vspace{8pt}
\subsubsection{Synthetic ID and Account Fraud} 
%\noindent \textit{Synthetic ID and Account Fraud}
Synthetic account fraud is based on fabricated identities made to look like real customers with favorable credit scores and characteristics. They regularly use social security numbers and credit privacy numbers blended with real and synthetic information from one or more individuals. Latest reports highlight that synthetic ID/account fraud has grown by around 35\% year-to-year \cite{Experian}. Synthetic ID fraud causes higher financial damage due to its non-transactional nature and the amount of time it takes to discover and report it.

%\vspace{8pt}
\subsubsection{Money Laundering} 
%\noindent \textit{Money Laundering}
Globally, between 715 billion to 1.87 trillion EUR is laundered according to the latest estimates \cite{Deloitte}. Money laundering aims to conceal the origin of the funds generated by criminal activity (such as drug trafficking and terrorist funding) to make it appear like the funds have originated from legitimate sources.  It typically involves \textit{layering}, during which multiple transfers occur between shell companies and individuals to conceal the sources.  Anti-money laundering \textit{(AML)}, know-your-client \textit{(KYC)} and countering financing of terrorism \textit{(CTF)} have been highly critical functions in financial institutions. However, anti-money laundering efforts have been facing serious difficulties in recent years.
%\vspace{8pt}

\subsubsection{Other Fraud Types}
Mortgage fraud and loan scams are quite prominent in the overall fraud landscape.  They generally involve customer account or personal identifying information (PII) compromises and often tie to identity theft. Hence, both the customers and the financial institutions suffer losses.  Financial services employees conducting fraud and criminal activities (such as bid rigging, price fixing, market allocation and bribery) is called internal fraud. Similar to external fraud detection, internal fraud solutions rely on entity-based interconnectivity and graph techniques to identify fraud rings.

%\subsubsection{General Characteristics}
Crime tactics display the ability to rapidly adapt to the emerging trends, vulnerabilities and prevention measures. They show remarkable levels of customization to the individual channels they operate on (as well as leveraging cross-channel tactics).  There are notable differences in fraud characteristics, in terms of transaction type (amounts, processing times), channel, devices, authentication requirements etc. As a result, ATM fraud is considerably different than online bill payment fraud (in terms of frequency, amounts, transaction and processing time ranges, parties involved, access compromises, devices involved etc.). These unique characteristics play a role in the effectiveness of the algorithmic solutions. However, such details are usually not taken into account in most research studies, which makes the implementation stages challenging.

%\vspace{8pt}

%%%%%%%%%%%%%%%%%%%%%%%%%%%%%%%%%%%%%%%%%%%%%%%%%%%%%
\subsection{Graph-based Detection Techniques}
\label{sec:Techniques}

\textit{Traditional AI and Machine Learning}
Machine learning has been used in fraud detection since the 90s \cite{Heaton16}, \cite{Chan98,Ghosh94,Aleskerov97}. Over the years, graph and network analysis techniques have been established as important tools in both research and industrial practice \cite{Mckinsey}.  Graphs inherently exhibit advantages in representing the underlying financial transaction data. The nodes and edges often represent companies, individuals, accounts, transfer of funds, locations, devices, and other financial or non-financial data. Depending on the application, a diverse range of graph types (directed, undirected, cyclical, acyclical, static, dynamic, attributed and colored) have been effectively utilized. 

\subsubsection{Data Mining}
\noindent Early data mining techniques for fraud detection relied on neural networks, regression, support vector machines, Bayesian networks that operated on tabular representations \cite{Ghosh94, Kim2003, Foster, Ezawa}. Later, graph-based representations have been explored \cite{chau,DataMining} such as community of interest selection \cite{Cortes01_2}, dynamic graphs \cite{Cortes03}, and signature-based systems \cite{Cortes01}.

\subsubsection{Graph Anomaly Detection} 
\noindent Anomaly detection has been effectively used in capturing the differentiating characteristics of fraud in the immense quantities of financial transaction data \cite{Akoglu2}, \cite{Fraud_Graph_Survey}. It provides insights on the data patterns by focusing on the distinct characteristics in terms of connectivity, flow, traffic patterns (both locally and globally) in the graph representations \cite{Akoglu,Noble}. 

\subsubsection{Supervised and Unsupervised Sub-Graph Analysis}
\noindent  Sub-graph analysis and mining analyzes the local graph patterns through supervised as well as semi- or unsupervised learning \cite{Faloutsos}, \cite{IBM_Molloy}. In addition to the connectivity, traditional network flow techniques (e.g. min-cut/max-flow) help identify the sub-graphs of interest. As an example, sub-graphs analysis may identify small-scale fraud rings generating abnormal patterns over a large number of accounts. Nevertheless, as fraud patterns are highly dynamic and adversarial, the use of a fraud detection technique causes changes in the fraud tactics to prevent detection. For instance, the use of strongly connected components by graph-based detection algorithms has motivated perpetrators to hide their activities by artificially creating networks to camouflage their activities. The use of metric-based analysis to detect such cases has been explored by some researchers  \cite{Faloutsos}. This highlights the inherent difficulties of fraud detection due to its adversarial nature.

\subsubsection{Flow and Path Analysis}
\noindent  In money laundering, the process of \textit{layering} directs the flow of illegal funds through multiple parties to prevent detection.  One of the limitations of the dense sub-graph based techniques is that they mostly focus on single-step transfers. Hence, they face difficulties and require adjustments to detect money laundering cases. Recently, multi-partite and multi-step solutions \cite{Flowscope}, flow analysis and k-step neighborhood based techniques \cite{Savage} have been proposed to address this. Network flow solutions, traditionally used for intrusion detection are also of great interest for financial crime detection use cases. 

\subsubsection{Graph-based Machine Learning}
\noindent In recent years, graph-based ML and graph neural networks have gained significant interest \cite{Scarselli2,Scarselli3}. Graph neural networks provide advantages over traditional neural networks in providing better generalizations and improved relational reasoning. Their applications in financial crime detection have yielded promising results \cite{GraphML}, \cite{GraphAdvantages}. Despite being in the early stages of exploration in fraud detection, GNNs provide flexible representations both in terms of attributes and graph structures. They enable a level of structural configurability along with the ability to compose architectures containing multiple blocks. Graph neural networks can operate on a variety of graph types.  Lately, numerous studies demonstrated GNNs in financial crime detection. \cite{weber} used graph convolutional networks (GCN) to detect money laundering. \cite{EvolveGCN} explored dynamic considerations in graph networks, which is essential in fraud applications. \cite{GNN_Camuflage} proposed solutions to expose camouflaged fraudsters through GNNs. Later studies aimed to improve the structural limitations of the graph networks through attention mechanisms for higher efficiency \cite{Graph_Attention}.

%%%%%%%%%%%%%%%%%%%%%%%%%%%%%%%%%%%%%%%%%%%%%%%%%%%%%
\section{Application Considerations}
\label{sec:Considerations}
This section discusses some of the common pain points and practical implementation considerations from a model development perspective. 

%\subsection{Temporal Considerations}
\subsubsection{Real-Time Processing and Response Time}
%\noindent\textbf{Real-Time Processing \& Response Time}
A large portion of the financial crime and fraud detection solutions is implemented within transaction processing systems. These complex systems process large amounts of transactional data in real-time. Transaction processing systems typically have millisecond-range response time SLAs. This end-to-end time constraint includes the transaction processing itself (accessing the account information, checking the availability of the funds), fraud scoring, payment network processing, communication protocols etc.  Graph computing solutions face serious response time pressures due to the combination of real-time processing constraints and the use of large, interconnected graphs. In the past decade, traditional solutions relied on memory efficient graph representations, graph compression techniques  \cite{shun} \cite{weber} and distributed graph computing to deal with the large graphs in financial use cases \cite{LargeGraph},\cite{DistributedGraph}. Similarly, graph attention networks eliminates the requirement of knowing the graph structure in advance and enables focusing on the most relevant parts of the input \cite{Graph_Attention}.  Fraud applications frequently borrow techniques from intrusion detection and cybersecurity in response to the agility pressures. Optimizing the system for data updates, selecting specialized temporal features, cost sensitive learning have been explored from intrusion detection solutions \cite{Stolfo, Stolfo2},\cite{Turley}. Other studies adapt the graph convolutional network (GCN) models along the temporal dimension to deal with dynamic complexities.

\subsubsection{Dynamic Graphs and Data Updates}
%\noindent\textbf{Dynamic Graphs \& Algorithm Selection}
Globally, medium to large banks serve tens of millions of clients on a daily basis over multiple processing channels. Payments channels are interconnected through entities (such as clients and accounts), hence the transaction throughput translates to tens of millions of updates to the underlying graph representations. High-frequency transaction types (e.g. P2P payments and credit card payments) require more frequent updates than slower channels (e.g. bill payments). Though they are in early stages, later techniques focus on the dynamic nature of the transaction systems through dynamic graph neural networks \cite{DGNN_Survey}. Evolving graph neural networks \cite{EvolveGCN}, streaming GNNs \cite{StreamingGNN}, temporal networks \cite{Temporal} have been proposed to perform deep learning on dynamic graphs. These solutions target a range of link durations and update frequencies.

%While high frequency payment channels require more research to adapt the emerging dynamic graph solutions, use cases such as synthetic fraud detection fare better in this regard due to their relatively static nature.

\subsubsection{Scale and Complexity Issues}
Financial transaction processing systems are typically complex, involving numerous payment transaction types and models to process fraud risk and business strategies.

%\vspace{2pt}
\textit{Number and Variety of Graphs:}
In financial payment systems the features of interest for fraud detection models may reside in different reference graphs of different types and disparate characteristics, internally and from external sources. This complexity becomes an obstacle in developing effective graph solutions. Due to their parallel structures, graph neural networks have parallelism potential, which, in theory, makes it possible to perform batch computations from independent graphs. Recently, techniques have been proposed to provide the ability to operate on diverse graph types \cite{Graph_Attention}.  Likewise, cross-channel fraud requires computations on multiple graphs with different characteristics simultaneously under the pressures of real-time response service-level agreements (SLAs). Though limited, some solution approaches have been proposed to integrate multiple graphs on unified graphs \cite{pena}, perform label propagation  \cite{Propagation},\cite{Attributed}, yet no clear solution has emerged to address the practical challenge.
%\vspace{2pt}

\textit{Entity and Feature-level Complexity:}
Financial transaction data usually includes the parties involved in the transactions, flow of funds, historical connectivity between parties, cross channel events, risk profiles etc. Depending on the use case, a large number of complex features may be used in graph-based techniques, these features need to be calculated in real-time to reach approve/deny decisions for the transaction.
\vspace{2pt}

\textit{Algorithmic Complexity and Response Time:}
While not emphasized in most studies, algorithmic complexity is also of interest for large scale and time-sensitive implementations such as fraud.  Graph neural networks have been shown to be effective on much larger networks than what they have been trained on, which is an advantage in large scale implementations \cite{Bello}. Still, the combination of size and algorithmic complexity drives serious implementation challenges. 

 %Furthermore, hybrid data sources and the overall data complexity pose challenges, where diverse set of data types may be needed such as unstructured data (text, images, video), entity graphs, transaction data.  

\subsubsection{Data Quality, Complexity and Noise}
%\noindent \textbf{Data Quality \& Noise}
In addition to the naturally occurring noise in the data, criminals frequently attempt to inject deceptive data into the financial transaction processing systems to conceal their activities. In synthetic account fraud, account takeover and other crimes, such data may become a determining factor in the decision making processes. While there are mitigation strategies for such attacks, robustness and performance of the underlying algorithmic solutions under data quality pressures are rarely considered in research studies.  
Financial crime detection often uses diverse data types and sources like unstructured data from images, video, audio recordings, text, emails, structured data from financial transactions, account information and historical records. However, continuous data integration and update processes, compounded with the data quality and complexity issues create a demanding environment and require targeted solutions.

\subsubsection{Labeled Data Availability and Infrastructure Support}
%\noindent \textbf{Labeled Data Availability}

%%%[H] I think we can mention sometimes entities relationship is difficult to discover and more susceptible to attacks. Although Graph NN can learn such relationships from the collected data to construct the adjacency matrix, how to improve its robustness remains a challenge in practice.
%Not sure..what do you mean by entity relationship discovery and susceptibility (?)
While most transaction-based detection systems (such as card payments) have well-established and automated infrastructure support for labeled data generation, other use cases may require special effort to generate such data.  Furthermore, labeled data limitations may also be caused by low fraud rates, lack of reporting or the timely integration of the data. Even though anomaly detection solutions exhibit natural advantages in cases with limited labeled data, improving the performance of unsupervised and semi-supervised solutions has been challenging.

\subsubsection{Adversarial and Robustness Issues} 
Fraud and financial crime detection are classic examples of adversarial applications. This adaptiveness can be (i) \textit{Response to the Prevention Measures:} As discussed in \Cref{sec:intro}, rule-based fraud detection approaches have largely become ineffectively as perpetrators quickly estimate the static rules and develop techniques to by-pass them. (ii) \textit{Shifts in Channel Usage and Schemes:} criminals have been utilizing emerging channels like bitcoin transactions for money laundering activities \cite{Bitcoin},\cite{worldbank_P2P}, fundamentally changing their tactics and adapting to the channel characteristics. (iii) \textit{Rapidly Changing Tactics over Time:} Similarly, mobile person-to-person payments are known to experience abrupt changes in fraud schemes, which makes supervised learning techniques challenging. Anomaly detection, and adaptive algorithms have been explored for these use cases. More importantly, adversarial tactics play a critical role in the overall robustness of the detection solutions. Often, the criminal organizations have the ability to react to prevention measures as well as the ability to scale their responses. In practical systems, this can have serious implications on the robustness of the machine learning models.

\subsubsection{Machine Learning Attacks}
Machine learning attacks are emerging concerns as they target highly-critical applications. Number of studies analyzed model robustness during possible poisoning, evasion, and inference attacks \cite{KDD}, \cite{Poison}, \cite{Robust}. Neural networks are known to be sensitive to even small perturbations in the data \cite{blackbox} and adversarial attacks. Hence, designing robust models has become a major implementation goal \cite{Szegedy, Goodfellow}. At this point, the research on robustness and adversarial issues for graph-based machine learning systems is in its nascent stages \cite{Adversarial}. 

\subsubsection{Governance and Regulatory Compliance}
Reinforcement learning provides key advantages in adversarial environments \cite{R_learning} thanks to its origins in gaming. Graph reinforcement learning studies have shown promising results in financial crime detection use cases \cite{reinforcement, Reinforcement19}. Yet, their implementation in the financial services industry has been slower. Similarly, autonomous learning approaches like online algorithms are of great interest for fraud detection \cite{deepwalk}. Nevertheless, the underlying autonomy and the lack of performance guarantees in these  techniques become issues for the model governance and regulatory compliance processes.  Current model governance processes require revisions and adjustments to speed up the adoption of more autonomous machine learning approaches for fraud detection.

\subsubsection{Interpretability and Explainability}
Regulations like CCPA \cite{CCPA} and GDPR \cite{GDPR} impose interpretability and explainability mandates on most regulated AI/ML models. Expanding the capabilities of the current explainability techniques is an industry-wide goal. Furthermore, developing specialized solutions that are optimized for emerging graph models and satisfy the regulatory requirements is highly critical \cite{Cvpr},\cite{Explainability2, LIME}.  

\subsubsection{Visualizations and Human-in-the-Loop Investigations}
Visualizations are used for multiple purposes in crime detection systems, such as in tracking emerging fraud trends, performing criminal investigations, investigating fraud alerts and collaborating with other fraud and financial crime analysts. Although they are quite effective in other fields, conventional visualization and human-in-the-loop tools face serious obstacles in fraud detection. Graph computing provides intrinsic advantages in representing and visualizing the data. Yet, scaling the solutions to meet the industrial application demands and dealing with the speed and complexity challenges are serious issues.

%%%%%%%%%%%%%%%%%%%%%%%%%%%%%%%%%%%%%%%%%%%%%%%%%%%%%%
\section{Conclusions and Outlook}
\label{sec:conclusion}

\noindent  Financial crime and fraud schemes have rapidly evolved in order to adapt to the new digital payments landscape. Graph-based solutions provide intrinsic advantages in detecting fraud in digital transaction data. Lately, graph neural networks provide promising results in a number of fraud detection use cases. However, implementing and deploying graph computing techniques in real-life detection systems pose unique difficulties.  These solutions face complications due to the size, speed, complexity and adversarial characteristics of the financial crime detection applications, which makes the deployments and reaching the detection performance targets difficult. In this paper, we overview the common application considerations and overarching implementation challenges graph-based solutions face in fraud and financial crime detection.

Graph neural networks and emerging adaptive solutions provide important opportunities to shape the future of fraud and financial crime detection. However, the complexity of the digital transaction processing systems (such as large-scale implementation requirements, real-time processing, multi-channel updates, complex data/graphs) and the ever-changing nature of fraud will likely continue posing challenges. The infrastructure and tool limitations require targeted efforts for these unique use cases. Moving forward, the adversarial tactics are likely to become greater challenges if robustness is not treated as a primary design goal in solution development.  Finally, focusing on the application demands and implementation issues have the opportunity to significantly improve the performance of current and emerging graph-based solutions.

\bibliography{bib}
\bibliographystyle{IEEEtran}
\end{document}